# TSMD: A Database for Static Color Mesh Quality Assessment Study

Qi Yang, Joel Jung, Haiqiang Wang, Xiaozhong Xu, and Shan Liu
*Tencent Media Lab*
{chinoyang, joeljung, walleewang, xiaozhongxu, shanl} @tencent.com

*Abstract*—Static meshes with texture map are widely used in modern industrial and manufacturing sectors, attracting considerable attention in the mesh compression community due to its huge amount of data. To facilitate the study of static mesh compression algorithm and objective quality metric, we create the Tencent – Static Mesh Dataset (TSMD) containing 42 reference meshes with rich visual characteristics. 210 distorted samples are generated by the lossy compression scheme developed for the Call for Proposals on polygonal static mesh coding, released on June 23 by the Alliance for Open Media Volumetric Visual Media group. Using processed video sequences, a large-scale, crowdsourcing-based, subjective experiment was conducted to collect subjective scores from 74 viewers. The dataset undergoes analysis to validate its sample diversity and Mean Opinion Scores (MOS) accuracy, establishing its heterogeneous nature and reliability. State-of-the-art objective metrics are evaluated on the new dataset. Pearson and Spearman correlations around 0.75 are reported, deviating from results typically observed on less heterogeneous datasets, demonstrating the need for further development of more robust metrics. The TSMD, including meshes, PVSs, bitstreams, and MOS, is made publicly available at the following location: https://multimedia.tencent.com/resources/tsmd.

## I. INTRODUCTION

A 3D mesh is a collection of vertices in 3D space, connected via edges that compose polygonal faces. It may have additional attributes associated with vertices, edges, faces, and corners, such as normal vectors and texture coordinates. Static meshes are widely used in many areas, such as animation, gaming, cultural heritage application, and medical imaging. It can provide vivid subjective perception and realistic use case simulation. It therefore attracts considerable attention in modern industrial design and manufacturing, such as rapid prototyping, digital fabrication, and precision measurement. In this paper, we focus on static meshes with texture map due to their common wide utilization in both academic and industrial fields.

Given the substantial volume of static mesh data, static mesh compression emerges as an extremely important and unavoidable technology. For instance, a static mesh with textures, that contains 30000 vertices and 100000 triangles, will typically have a raw data size of about 2.5GB. Therefore, an effective mesh compression algorithm can reduce the computational workload and resource cost (e.g., memory and bandwidth): for rendering, compression can improve performance and efficiency by loading fewer data on Graphics Processing Unit (GPU). For mobile devices and network environments, compression can reduce computing power and storage resources and save the cost of valuable bandwidth.

Current static mesh compression tools have weaknesses, which prevent optimal compression efficiency. They only quantize vertex attributes with a minimal loss of visual quality, targeting mainly GPU rendering [7]. The next-generation mesh compression standards would improve compression efficiency for all types of polygonal 3D meshes with more encoding strategies, such as decimation proposed in dynamic mesh compression [8], or spatial scalability [18]. Developing effective and mature mesh compression algorithms needs the assistance of automatic quality evaluation, via robust objective metrics. For lossy compression, rate-distortion (RD) curves are used to measure compression efficiency [24]. Given an expected quality level, a reliable quality evaluation can assist us in picking up a superior codec tool, selecting optimal compression configuration and saving bandwidth costs. More generally, quality metrics contribute to increasing customer satisfaction through better quality of experience, with controlled computation, bandwidth, and resource costs.

Objective quality evaluation for static mesh has been studied for nearly 25 years. From point-wise features (e.g., Hausdorff distance [9]) to structure-wise features (e.g., global smoothness [10], curvature [11]), model-based static mesh metrics gradually show better performance by taking more and more human visual system characteristics into consideration. However, these metrics still have some limitations: most of them only consider geometry features while the texture features are ignored. They also have strict requirements on distorted samples, such as sharing the same connectivity, the same vertex density, or the same level of detail [12]. MPEG Coding of 3D Graphics and Haptics working group (WG7) proposes to consider image-based and point-based metrics, in addition to model-based metrics [12]. Although these types of metrics have the potential to compensate for some deficiencies of model-based metrics, they have limitations: 1- the preprocessing (projection, sampling) adds computation time overhead. Image-based metrics use multiple viewpoints, and point-based metrics require sampling to generate points face-by-face, 2- the Fibonacci sphere lattice used by WG7 [25] to select viewpoints for image-based metrics, adequate for most meshes, can produce viewpoints unlikely to be observed by a human, for some meshes representing indoor scenes. Besides, the projected images may contain background information, which can impact the measure of the mesh distortion. For point-based metrics, the performance varies with the sampling methods and sampling resolutions.

New datasets with diverse content, varying levels of degradation, and accurate mean opinion scores (MOS) are needed for several reasons. First, a dataset is needed to design efficient objective metrics [2][3]. Testing the performance of objective metrics on a complete dataset helps to judge whether the metrics correlate well with MOS and are robust to multiple kinds of content and degradations. Second, an effective objective metric can facilitate the development of compression algorithms: it can contribute to optimal encoder decisions with RD criterion, and help with decision taking on tool adoption. Therefore, a dataset is also necessary for compression algorithm study.

As far as we know, there are only two datasets for static mesh including texture map. [1] contains five references and five types of distortions: the diversity of this dataset is reduced. Only a single distortion is introduced, while compression algorithm will typically introduce several degradations, due to cascaded encoding modules. [2] proposes 55 references with quantization, texture map sub-scaling, and texture map compression applied successively. Considering the availability of only two datasets, it is necessary to construct a new one that provides diverse samples and trustworthy MOS. It will facilitate research on mesh compression and quality assessment.

In this paper, we establish a new dataset for static mesh with texture map, called Tencent – Static Mesh Dataset (TSMD). It includes 42 reference meshes with rich visual characteristics from different sources: human, inanimate object, animated graphic, animal, indoor scene, plan, and building mesh samples. Four types of distortions are successively introduced to generate distorted meshes with five different degradation levels, yielding 210 distorted samples. Processed video sequences (PVSs) were generated for subjective experiments with a crowdsourcing environment. The mesh samples, rate information and PVSs will be publicly released.

The remainder of this paper is organized as follows. Section II details the construction of the dataset. Section III analyzes the dataset in terms of sample diversity and MOS accuracy. Section IV reports the performance of famous objective metrics on the new dataset. Section V summarizes the whole paper and envisions future work.

## II. Dataset Creation

This section presents comprehensive information regarding the content of the dataset and the methodology employed to derive the subjective scores.

### A. Content description

We have compiled 42 publicly available meshes from reputable sources: Volograms&V-SENSE [4], VoluCap[5] and SketchFab [6]. They encompass a wide range of static content, applicable to various virtual reality use cases, such as human characters, inanimate objects, indoor scenes, building, etc. The essential information of the dataset is summarized in Table 1. To illustrate the diversity of content, Fig.1 provides a snapshot of a selection of meshes.

Table 1: Summary information of the TSMD.

| | |
|---|---|
| Number of meshes | 42 |
| Distortions applied to the mesh (successively) | Dequantization, Triangulation, Decimation, Draco compression |
| Distortions applied to the texture (successively) | Downscaling, AV1 (libaom) compression |
| Levels of distortion | 5 |
| Total number of distorted meshes | 210 |
| Total number of scores collected via crowdsourcing / remaining scores after outlier removal | 10320 / 9468 |

### B. Distortion generation

Four distinct types of distortion are introduced successively to the source content. They correspond to the distortions applied in the context of the VVM polygonal static mesh coding Call for Proposals (CFP) [18] of AOMedia, as depicted in Fig. 2a. Each mesh is successively dequantized, triangulated, decimated, encoded and decoded by Draco. Each texture undergoes first a downscaling, to meet the following constraints: the vertical/horizontal ratio is kept unchanged, none of the horizontal or vertical resolution exceeds 4096 pixels. Then AV1 compression is applied. For each distortion, five encoding configurations are carefully selected, made of the following parameters: decimation level, Draco quantization parameter for position attribute and for the texture attribute (normals are not quantized), and AV1 quantization. This allows us to reach five rate-distortions points, evenly distributed over practical quality ranges applicable to various use cases.

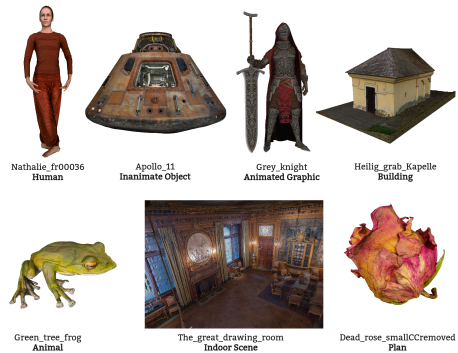

Fig 1. Snapshots of a selection of meshes.

### C. PVS generation

Following a conventional approach, Processed Video Sequences (PVS) are generated from both source and distorted content. These PVS are 2D videos that depict camera paths simulating typical user motions. They are displayed during subjective tests to enable the evaluation process. The initial step involves rendering the content from the distorted meshes and textures, utilizing the open3D library. The object is centered, and its size is adjusted to occupy a significant portion of the frame. A rotational movement is applied to the object, following a cosine function that gradually decreases the rotation speed until near stop, and then increases it again. Two meshes have considerably distinct characteristics compared to the others, as they represent scenes inside a building. Consequently, for these meshes, the camera is positioned inside the scene before the rotational movement is applied.

The resulting PVS are created with a frame rate of 30 fps and a duration of 18 seconds. These PVS are subsequently encoded using the FFMPEG library, with the x264 encoder and a constant rate factor equal to 10 to ensure visually lossless encoding. This approach aligns with the guidelines proposed in [19].

### D. Subjective experiment

To conduct the subjective tests, a crowdsourcing methodology is implemented, utilizing a proprietary crowdsourcing interface. It enables the download and streaming of the PVS to the participants, who are then prompted to provide ratings based on the displayed content.

Before the rating session starts, a comprehensive set of instructions is provided to the participants to ensure the clarity and consistency of the test. Of utmost importance, a training session is conducted, which closely replicates the rating session. In the training session, participants are required to assign scores within the range of "possible" scores for 8 carefully selected PVS, excluding any unrealistic scores. Successful completion of the training session serves as a qualification criterion for participants to proceed to the actual rating session.

The complete methodology follows ITU-T P.910 Recommendation [20] yet is adapted to the crowdsourcing approach. A Double Stimulus Impairment Scale is used. The source content and distorted content are shown in pairs, in random order, before the score is requested, using a five-level scale for rating the impairments.

The participants are so-called "naïve" viewers, not familiar with mesh related research. They comprise a group of 74 viewers of age between 17 and 25, consisting of students.

*E. Outlier detection*

In the context of crowdsourcing, the detection and removal of outliers play a vital role. To tackle this issue, specific "trapping" PVS are intentionally included within the rating sessions. These PVS correspond 1- to extremely low quality PVS for which extremely low scores are expected, and 2- to duplicated PVS for which close scores are expected. By incorporating these trapping PVS, it becomes possible to identify and filter out any outliers during the evaluation process.

Furthermore, the correlation between the average score and each group of raw data is calculated. If the correlation value falls below 0.8, the corresponding data is deemed unreliable and subsequently excluded from the dataset. Through this process, 852 scores were removed from a total of 10320. It is guaranteed that each PVS takes advantage of a remaining number of scores above 15 to compute the Mean Opinion Score (MOS), once outliers are rejected.

### III. DATASET VALIDATION

In this section, we demonstrate that the proposed dataset provides samples with rich visual characteristics and reliable MOS.

*A. Diversity of content*

To measure the diversity of the dataset content, spatial information (SI) and temporal information (TI) are calculated based on the PVSs viewed in subjective experiments. SI takes texture structure into consideration, and TI measures motion differences. Both indicators reveal the difference in geometry and texture characteristics of 3D meshes under different viewpoints. SI vs TI plot is illustrated in Fig. 2b. TSMD covers a wide range of SI and TI values. The indoor scene "great_drawing_room" mesh has the largest SI value, and the largest TI value is for the "butterflies_collection" mesh. The inclusion of indoor scene meshes within the dataset significantly enhances its overall diversity.

*B. Accuracy of MOS*

To verify the MOS accuracy of the dataset, we examine the correlations between MOS and bitrate. For a given mesh, a higher bitrate is supposed to indicate more detailed visual features, so higher MOS ratings. Fig. 3 illustrates the bitrate vs MOS for two meshes. The curves show perfect monotony.

Some exceptions occur, such as for the "grey_knight" mesh. When the decimation rate increases from 0.1 to 0.6 (more faces and vertices are retained), the MOS does not reflect the expected quality improvement. The change of decimation level does not generate enough visual impact. An expert viewing of the corresponding PVS confirms that the MOS obtained is valid: indeed, the bits were not optimally used by the compression. The "grey_knight" mesh exhibits relatively low visual information (SI=0.015, TI=0.032). The presence of predominantly black and dark red textures partially masks distortion. This is why the decimation rates of 0.6 and 0.1 appear similar. Perfect bitrate vs MOS is achieved for the majority of meshes and confirms the reliability of our subjective experiment.

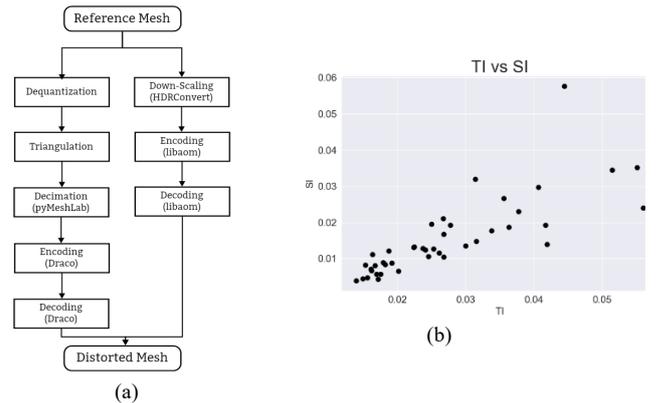

Fig 2. (a). Static mesh compression anchor[18]; (b) SI vs TI.

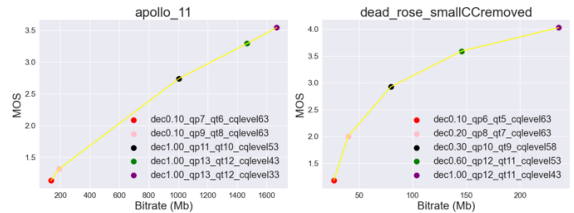

Fig.3. Example of monotonic bitrate vs MOS curves.

### IV. PERFORMANCE OF OBJECTIVE METRICS

We have confirmed that TSMD is a reliable dataset with heterogeneous content. It is therefore appropriate for evaluating the efficiency and robustness of objective metrics. There are two major questions: what is the performance of the state-of-the-art metrics on the new dataset, and are some of them robust enough to report high correlation on a heterogeneous dataset?

Three types of metrics are tested: image-based (geo-PSNR and YUV-PSNR), point-based (point-to-point D1, point-to-plan D2 [17], and PCQM-PSNR [16]), and video-based (PSNR, SSIM [14], and VMAF [15]). Image-based and point-based metrics are calculated via MPEG-pcc-mmetric software [12]: the resolution of projected images is 1920x1920, grid sampling is applied with 1024 resolution. Video-based metrics are calculated with the MSU Video Quality Measurement Tool [13] on the PVSs used for the subjective experiments.

*A. Correlation of objective metric*

After mapping the objective scores to a common scale using a five-parameter logistic regression [21], we report PLCC, SRCC, and RMSE results on the whole dataset in columns "All" of Table 2. PCQM-PSNR reports the best performance, followed by geo-PSNR, D1 and VMAF. Fig. 4 shows scatter plots of PCQM-PSNR, and VMAF. Both have many points away from the best fitted logistic regression curve, represented in yellow. The best overall performance, around 0.75, is far from the usual reported correlation and from what characterizes a robust metric (above 0.85). This weakness is demonstrated thanks to the diversity of TSMD.

*B. Analysis by category of content*

For a more in-depth analysis of the performance of the objective metrics, the PLCC is reported in "AOMedia Classes" columns of Table 2 for six different subclasses, defined in AOMedia VVM CfP [18]. A2 collects game characters obtained by digital content creation (DCC), B is a collection of avatars captured as 3D scans (3DS) of humans, D consists of professional 3DS (D-1 with only triangular faces and D-2 with any polygonal faces), E contains non-professional 3DS of objects, and F professional 3DS from outdoor and indoor scenes. We also split the dataset based on the mesh creation process, DCC or 3DS, and report the results in the Table 2 "Creation" columns.

It can be observed that PCQM-PSNR does not report the best results for all contents and conditions. It is the best on classes D-1, D-2, F, and 3DS. Some video-based metrics show the best results on class A2 and E, with correlations above 0.90, but fail completely on classes B and F. D1 and D2 metrics exhibit impressive results on class B. Most metrics report close performance between classes D-1 and D-2. For DCC class, image-based metrics exhibit good performance, and the gap between the different types of metrics is relatively small. For 3DS class, metrics report poorer performance than for class DCC.

Table 2. Metrics performance on TSMD

|   | Metric | All | | | AOMedia Classes | | | | | | Creation | |
|---|---|---|---|---|---|---|---|---|---|---|---|---|
|   |        | PLCC | SRCC | RMSE | A2 | B | D-1 | D-2 | E | F | DCC | 3DS |
| I | geo-PSNR | 0.73 | 0.73 | 0.80 | 0.92 | 0.90 | 0.72 | 0.71 | 0.87 | 0.74 | **0.93** | 0.73 |
| I | YUV-PSNR | 0.68 | 0.68 | 0.85 | 0.85 | 0.83 | 0.76 | 0.88 | 0.76 | 0.80 | 0.86 | 0.72 |
| II | D1 | 0.72 | 0.65 | 0.80 | 0.93 | **0.95** | 0.73 | 0.68 | 0.65 | 0.77 | 0.92 | 0.71 |
| II | D2 | 0.54 | 0.47 | 0.98 | 0.84 | 0.93 | 0.69 | 0.52 | 0.48 | 0.44 | 0.85 | 0.54 |
| II | PCQM-PSNR | **0.76** | **0.75** | **0.76** | 0.85 | 0.85 | **0.81** | **0.91** | 0.76 | **0.89** | 0.82 | **0.78** |
| III | PSNR | 0.53 | 0.54 | 0.99 | **0.95** | 0.55 | 0.66 | 0.65 | 0.80 | 0.44 | 0.86 | 0.55 |
| III | SSIM | 0.51 | 0.67 | 0.85 | 0.78 | 0.53 | 0.64 | 0.63 | 0.74 | 0.52 | 0.74 | 0.48 |
| III | VMAF | 0.70 | 0.51 | 1.00 | **0.94** | 0.63 | 0.79 | 0.76 | **0.91** | 0.61 | 0.87 | 0.69 |

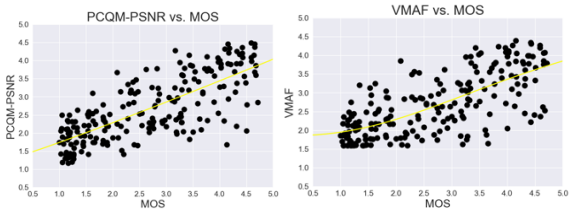

Fig 4. Scatter plot of PCQM-PSNR and VMAF vs MOS.

*C. Extended analysis of objective metric results*

Some metrics present inconsistent performance across different subclasses. The "bitrate vs objective scores" plots indicate that one reason for poor metrics' performance is that different scoring ranges are produced for samples that share close MOS. The reason why metrics give "inconsistent" ranges of objective scores depends on the type of metric.

For video-based metrics, one weakness is that scores of indoor meshes are generally lower than scores of other meshes. The proportion of background content, static and artifact-free, contained in the frame significantly affects the range of quality scores. The frame for indoor meshes has no such background because the inside of the building covers the full frame. It consequently has more distorted information counted by the metrics, leading to a smaller quality score. It is expected that applying the metrics on the bounding box of the meshes would help with this issue.

Image-based metrics suffer from the same weakness. In addition, the selection of the viewpoints based on MPEG WG7 approach results in non-realistic viewpoints for indoor scenes meshes. Some bias is consequently introduced for class F.

For point-based metrics, the performance is tight to the sampling of the point clouds. The impact of mesh distortion on sampling results is hard to predict, causing unstable performance. For instance, all point-based metrics present poor performance on the "japanese_spiny_lobster" mesh. This mesh has a very complex geometrical structure, and for grid sampling, the number of sampled points based on distorted mesh does not present monotonic variation with the increase of bitrate. Generally, such metrics are sensitive to the variation of density of points, between the reference and the distorted point clouds [16][22][23]. Some point-based metrics even assume that point clouds with more points should have a better quality. The cases in which distorted meshes generate more points than the reference mesh can cause the unstable results of point-based metrics.

## V. CONCLUSION AND FUTURE WORK

This paper introduces the development and the public release of a novel static mesh dataset. It consists of 42 static textured meshes with diverse content and varying levels of degradation, along with subjective scores obtained through crowdsourcing. The availability of this dataset holds significant importance for both the academic research community and the industry. It enables the design of objective metrics and facilitates the development of mesh compression techniques. To the best of our knowledge, it offers the most comprehensive and diverse range of samples. It can serve as an essential resource that enables a thorough assessment of the robustness of metrics across various types of content.

The dataset undergoes an analysis to validate its sample diversity, establishing its heterogeneous nature. The reliability of the scores is demonstrated through MOS vs bitrate plots. The new dataset is used to evaluate three types of state-of-the-art objective metrics. Previous research often reports high PLCC and SRCC correlations (around 0.9) for the top-performing metrics. However, it is demonstrated that when utilizing a heterogeneous dataset, the best metrics only achieve a correlation around 0.75. Furthermore, the metrics exhibit unstable performance across sample categories. For instance, video-based metrics struggle with indoor scene meshes. This dataset serves as a catalyst for new research endeavors on the robustness of objective metrics, providing opportunities for further investigations.

Among the future works, we aim to conduct subjective experiments within a virtual reality environment, utilizing the content available in this dataset, and we strive to enhance the robustness of objective metrics.